\documentclass[10pt,twocolumn,letterpaper]{article}

\usepackage{wacv}
\usepackage{times}
\usepackage{epsfig}
\usepackage{graphicx}
\usepackage{amsmath}
\usepackage{amssymb}
\usepackage{import}
\usepackage[utf8]{inputenc}
\usepackage{fourier} 
\usepackage{array}
\usepackage{makecell}
\usepackage{tabularx}
\usepackage{multirow}
\usepackage{multicol}
\usepackage{float}
\usepackage{xcolor}
\usepackage[skip=10pt]{caption}
\newcommand{\rev}[1]{{\color{black} #1}}

\makeatletter
\@namedef{ver@everyshi.sty}{}
\makeatother
\usepackage{tikz}
\usepackage{textcomp}
\usepackage{lipsum}

\newcommand\copyrighttext{%
  \footnotesize \textcopyright This work has been accepted for publication in the IEEE/CVF Winter Conference on Applications of Computer Vision 2023. Copyright with IEEE. Personal use of this material is permitted. However, permission to reprint/republish this material for advertising or promotional purposes or for creating new collective works for resale or redistribution to servers or lists or to reuse any copyrighted component of this work in other works must be obtained from the IEEE. For more details, see the IEEE Copyright Policy.}
\newcommand\copyrightnotice{%
\begin{tikzpicture}[remember picture,overlay]
\node[anchor=south,yshift=10pt, xshift=-9pt] at (current page.south) {\fbox{\parbox{\dimexpr\textwidth-\fboxsep-\fboxrule\relax}{\copyrighttext}}};
\end{tikzpicture}%
}

\usepackage{pict2e}

\DeclareRobustCommand{\diam}{%
  \begingroup
  \setlength{\unitlength}{\fontcharht\font`T}%
  \begin{picture}(1,1)
  \polygon(.6,0)(1.2,.6)(.6,1.2)(0,.6)
  \end{picture}%
  \endgroup
}
\DeclareUnicodeCharacter{25C8}{\diam}
\newcommand{\dplus}{\,$\diam$\kern-0.68em\raisebox{1.1pt}{$+$}\,\,}

\def\wacvPaperID{1142} 

\wacvapplicationstrack 

\wacvfinalcopy 

\ifwacvfinal
\usepackage[breaklinks=true,bookmarks=false]{hyperref}
\else
\usepackage[pagebackref=true,breaklinks=true,colorlinks,bookmarks=false]{hyperref}
\fi

\begin{document}

\title{DeformIrisNet: An Identity-Preserving Model of Iris Texture Deformation}

\author{Siamul Karim Khan, Patrick Tinsley, Adam Czajka\\
University of Notre Dame, IN, USA\\
{\tt\small \{skhan22, ptinsley, aczajka\}@nd.edu}
}

\maketitle
\copyrightnotice
\thispagestyle{empty}


\begin{abstract}
Nonlinear iris texture deformations due to pupil size variations are one of the main factors responsible for within-class variance of genuine comparison scores in iris recognition. In dominant approaches to iris recognition, the size of a ring-shaped iris region is linearly scaled to a canonical rectangle, used further in encoding and matching. However, the biological complexity of the iris sphincter and dilator muscles causes the movements of iris features to be nonlinear in a function of pupil size, and not solely organized along radial paths. Alternatively to the existing theoretical models based on the biomechanics of iris musculature, in this paper we propose a novel deep autoencoder-based model that can effectively learn complex movements of iris texture features directly from the data. The proposed model takes two inputs, (a) an ISO-compliant near-infrared iris image with initial pupil size, and (b) the binary mask defining the target shape of the iris. The model makes all the necessary nonlinear deformations to the iris texture to match the shape of the iris in an image (a) with the shape provided by the target mask (b). The identity-preservation component of the loss function helps the model in finding deformations that preserve identity and not only the visual realism of the generated samples. We also demonstrate two immediate applications of this model: better compensation for iris texture deformations in iris recognition algorithms, compared to linear models, and the creation of a generative algorithm that can aid human forensic examiners, who may need to compare iris images with a large difference in pupil dilation. We offer the source codes and model weights available along with this paper.
\end{abstract}

\section{Introduction}

The human iris is a thin, circular structure presenting a rich and unique texture information defined by detailed characteristics such as crypts, ridges, furrows, rings, corona, freckles, and a zigzag collarette. The minute iris patterns and textures are formed during gestation and have relatively little influence from genes \cite{daugman1993high}, making it unique 
even for the identical twins, and different in our left and right eyes. 
Due to its uniqueness and stability, iris texture is one of the most reliable biometric traits widely employed for recognition. Recent research results \cite{Trokielewicz_TIFS_2019} also suggest that iris recognition is feasible a few weeks after death, which opened interesting additional applications to forensic identification.


One of the primary sources of within-class variance in this biometric modality is the highly complex iris texture deformation due to pupil size variation. Dominant approaches to iris recognition based on Daugman's method \cite{daugman2009iris} map the iris annulus into a rectangular region of canonical dimensions. While simple and effective, this linear mapping does not compensate for large differences in pupil size~\cite{hollingsworth2009pupil}. This is why a nonlinear mapping, modeling some but not all of the possible iris texture deformation subtleties~\cite{tomeo2015biomechanical}, was found to be a better choice.

\begin{figure*}[!htbp]
\centering
\includegraphics[width=0.93\textwidth]{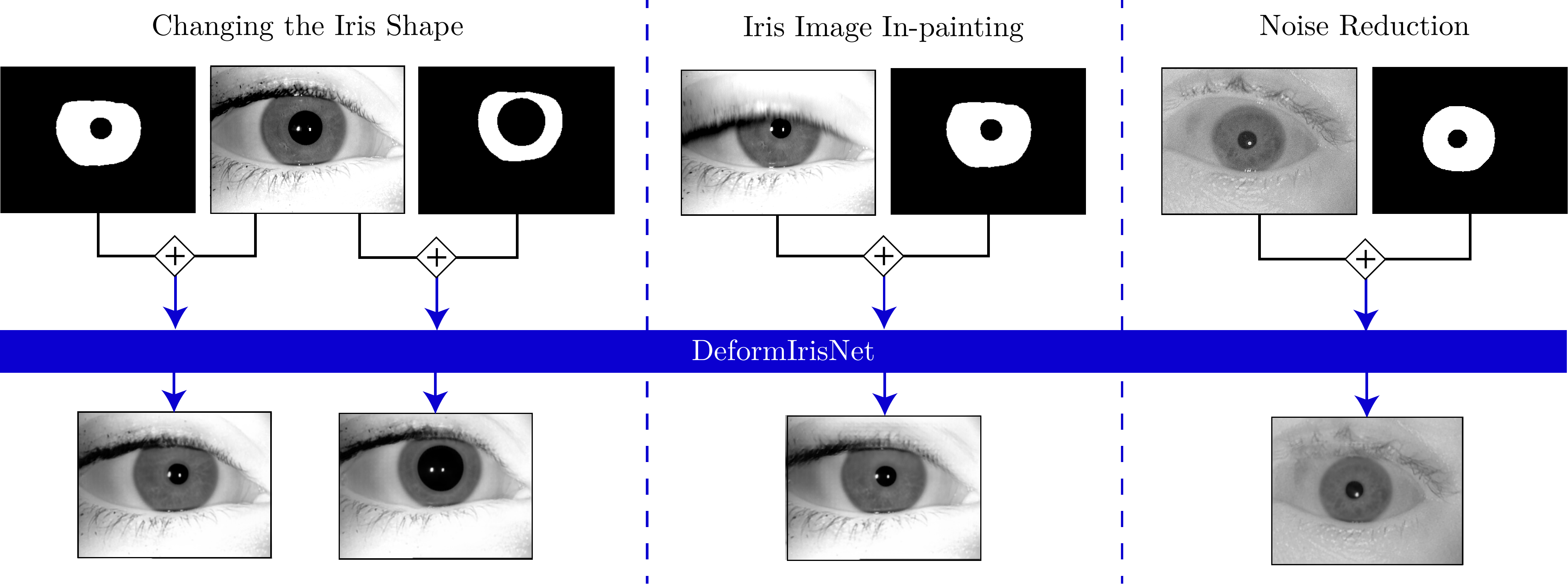}
\caption{The proposed model (DeformIrisNet) deforms an input iris image ({\bf left}) to match the requested shape of the iris given by the target mask (also provided as an input), preserving the identity and improving the iris recognition compared to linear normalization methods. Deformation is made in a way to mimic the authentic, very complex iris muscle movements. Additionally, the proposed model can ``inpaint'' iris texture in places covered by eyelids in the input sample ({\bf middle}) and reduce selected types of noise in the input image when we do not request the change in iris size ({\bf right}). \dplus denotes concatenation across channels.}
\label{fig:teaser}
\end{figure*}

\rev{In this paper, we propose an {\bf end-to-end deep-learning-based model of complex iris texture deformations}. We demonstrate its usefulness in iris recognition (to match the pupil size between iris scans being compared). We also propound that this method has huge potential to aid in human forensic iris examination.} The model does not assume circularity of iris boundaries and is able to process full-frame ISO-compliant \cite{ISO_19794_6_2011} iris images (not only their normalized versions).
In particular, we propose an autoencoder trained with identity-preserving loss that deforms an input iris image to match its new shape given by an input mask, as shown in Fig. \ref{fig:teaser}. The proposed method is trained to appropriately capture the complex, non-linear dynamics of the iris texture and generate iris images with dilated \rev{or constricted} pupil which can allow for better iris recognition performance. Our model can also aid human examiners in better comparing iris image pairs with excessive differences in pupil size, which was found to be one of the most difficult issues for subjects comparing iris images because the large iris texture deformations in these cases made the same salient features across images look quite different~\cite{Moreira_WACV_2019}. We also show that an intuitive approach employing components of modern Generative Adversarial Networks may offer results (in terms of identity preservation) that are inferior to those obtained with the proposed autoencoder-based model. \rev{The main {\bf contributions} of the paper are:
\begin{enumerate}
    \item An end-to-end, data-driven, deep learning-based model of complex iris texture deformations;
    \item An iris recognition-based identity-preserving loss component that aims at constraining the deformations to those mimicking the authentic iris muscle movements;
    \item An iris mask loss component that maintains the shape of the iris, which, together with the perceptual similarity loss can, additionally to (2), lead to realistically-looking iris images.
\end{enumerate}
}

We offer source codes and model weights along with this paper at \url{https://github.com/CVRL/DeformIrisNet}. 


\section{Related Works}

The most commonly used method for iris normalization is the homogeneous ``rubber sheet'' model proposed by Daugman~\cite{daugman1993high} which maps every pixel $(x, y)$ of the iris annulus in the Cartesian coordinate system into an equivalent pixel in the polar coordinate system $(r, \theta)$, where $r\in [0, 1]$ is the radial distance from the inner iris boundary, and $\theta \in [0, 2\pi]$ is the angular position. This model converts the circular iris region into a canonical rectangular image, compensating linearly for pupil size variations. Hollingsworth \etal have shown, however, that despite this linear normalization, pupil dilation can degrade iris recognition performance~\cite{hollingsworth2009pupil, bowyer2009factors}. 

There are relatively few advances in iris normalization compared to iris segmentation and iris feature extraction. Tomeo-Reyes \etal~\cite{tomeo2015biomechanical} addressed the problem of pupil dilation by proposing the biomechanical model of iris muscle deformation. They model the iris region as a thin cylindrical shell made of orthotropic material and utilize the biomechanics of the iris region to calculate the displacements that occur during iris dilation and constriction. This model can compensate for some, but not all nonlinear movements of iris features, as it is a significant simplification of the dynamics of the iris musculature. 
Wilde \etal~\cite{wildes1997iris} compensated the iris deformation using an image registration technique. The technique searches for an optimal transformation in both space and intensity that maps each point from one image to a point in the other. Wei \etal~\cite{wei2007nonlinear} proposed an alternative non-linear iris normalization model based on statistical learning. The
proposed iris deformation model is a combination of linear and non-linear stretch. The linear stretch is based on the rubber-sheet model and the non-linear stretch is modeled using a Gaussian function with parameters obtained by training. Yuan \etal developed a non-linear iris normalization method based on the minimum-wear-and-tear meshwork of the iris~\cite{yuan2005non}. The minimum-wear-and-tear meshwork proposed by
Wyatt~\cite{wyatt2000minimum} models the iris as a mesh of radial and circular muscles and estimates the deformation based on the motion of this meshwork. Lefevre \etal proposed a ``rubber sheet'' model based on fitted ellipses instead of circles~\cite{lefevre2013effective}. 
More recently, Generative Adversarial Networks-based approaches have also been experimented with to vary pupil size. However, most of such methods suffer from the ``texture sticking'' problem making identity preservation difficult. 

The model proposed in this paper {\bf is different from the previous work} in two ways. First, it makes neither geometrical nor biological assumptions about the iris constriction phenomenon, nor requires providing the ``pupil size'' (which is actually difficult to define due to the irregular shape of the pupil). The proposed model learns the complex iris muscle movement directly from the data. Second, the model works directly with ISO-compliant iris scans (\ie $640\times 480$ pixel near-infrared images), which makes it applicable to complement {\it any} iris recognition algorithm, including ``black box'' or closed commercial matchers. 



\section{Database}
We utilize the Warsaw-BioBase-Pupil-Dynamics v3.0 (WBPD) dataset which consists of 30-second near-infrared iris videos with variable pupil size due to visible-light stimuli~\cite{kinnison2019learning}. 
After 15 seconds, a visible light stimulus caused pupil contraction, which was kept for 5 seconds, and then followed by a restorative dilation. After converting original videos to images, the dataset contains 117,117 grayscale images at a resolution of $768\times576$ pixels, representing 84 different eyes. 

A necessary data curation step was to pair small and large pupil images in a way that allows us to provide appropriate inputs and targets to train the proposed model. To do that, we first automatically detect the pupil and iris radii for all the images in the dataset and find the pupil-to-iris ratio. Previous research has shown that the pupil-to-iris ratio generally varies between 0.2 (highly constricted pupil) and 0.7 (highly dilated pupil) ~\cite{hollingsworth2009pupil, tomeo2015biomechanical}. We take all the images with a pupil-to-iris ratio between 0.2 and 0.7, and divide them into 5 bins of width 0.1. That is, the first bin contains all images with pupil-to-iris ratio between 0.2 and 0.3, the second bin contains all images with pupil-to-iris ratio between 0.3 and 0.4, and so on. As we are concerned with training a model for dilation, for images in each bin we pair them with images from all the bins with a higher pupil-to-iris ratio. 

We divide the data into eye-disjoint training, validation, and test sets, with samples from 67 eyes in the training set, samples from another 6 eyes in the validation set, and remaining samples representing 7 eyes in the test set. This eye-disjoint split is crucial to make sure the model is learning generic iris muscle movements, and not dynamics specific to subjects present in the training dataset.

\section{Proposed Method}

\subsection{Overview}
We propose a modified U-Net autoencoder, which is trained to construct a new image with an iris matching the shape (including size) delivered by the image mask provided also as input. Though it is not the autoencoder architecture that is our contribution, and rather its training mechanism with multiple loss components, that ends up with a possibility to learn complex iris muscle movements to generate images that not only preserves the identity but even helps in iris recognition. \rev{As the input mask can be of any shape, our model can both dilate and constrict the pupil as necessary.}

\begin{figure}[!ht]
\centering
\includegraphics[width=\linewidth]{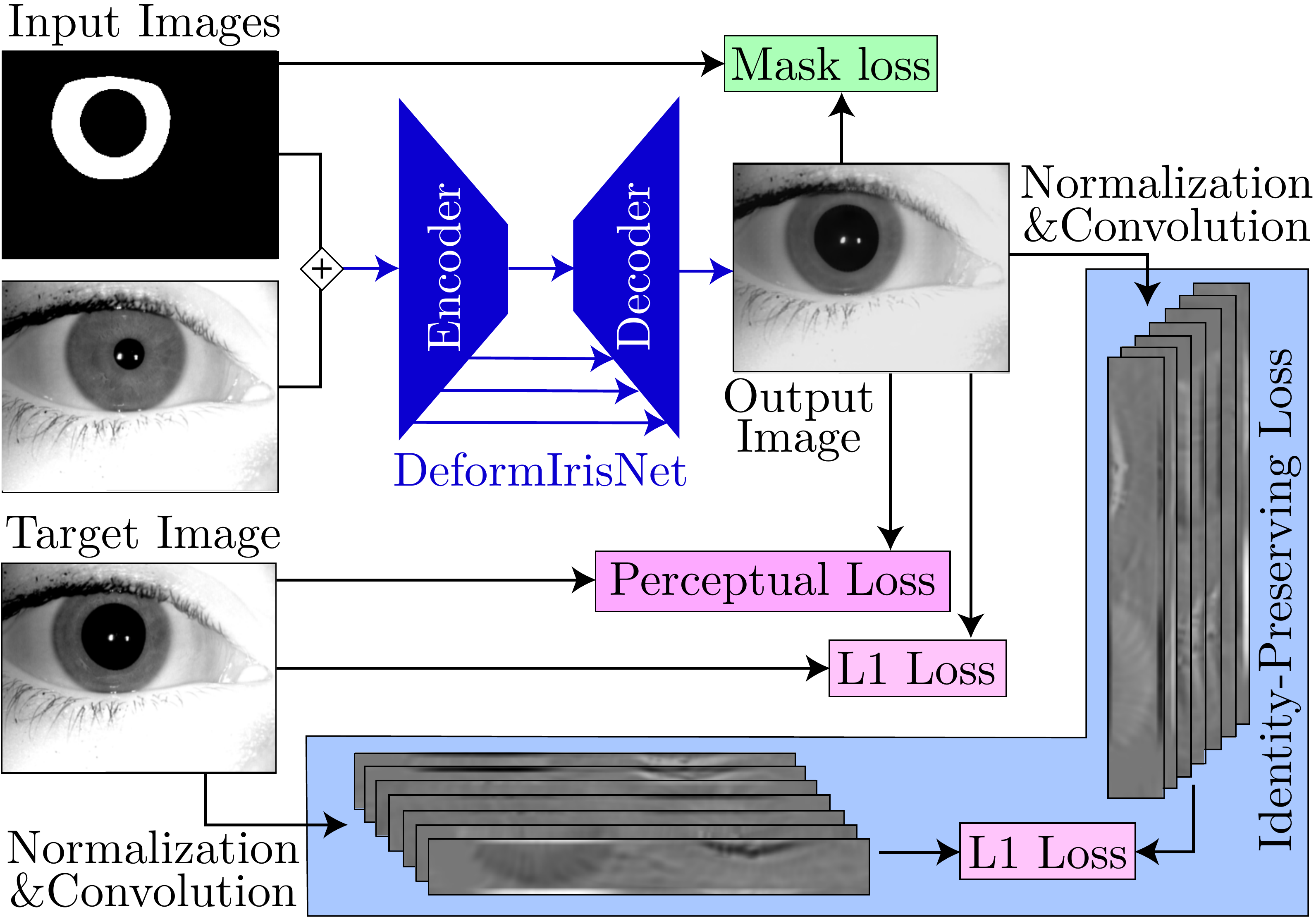}
\caption{Illustration of the training mechanism with all loss components: (a) ``Mask Loss'' to request new shape of the iris, (b) ``Identity-Preserving Loss'' for constraining the deformations to those mimicking the authentic iris muscle movements, hence preserving the identity information in the generated sample, (c) ``L1 Loss'' to maximize general similarity of the generated and ground-truth images, and (d) ``Perceptual Loss'' used to maximize perceptual realism of the generated samples. \dplus denotes concatenation across channels.}
\label{fig:train}
\end{figure}

Figure~\ref{fig:train} presents an overview of the model's training strategy. All components of the proposed loss function can be grouped into two functional sets: identity preservation component and perceptual realism preservation component. The next two subsections describe these two aspects of the training in detail.

\subsection{Identity Preservation}
To preserve the identity of the iris while training the autoencoder, we define a loss that specifically focuses on identity preservation. First, we convert the iris regions of both the output and the target from Cartesian coordinates to a pseudo-polar coordinate system \cite{daugman1993high}. For an image $I$ with iris center $(x_i, y_i)$, iris radius $r_i$, pupil center $(x_p, y_p)$ and pupil radius $r_p$, we can build a normalized iris image as a function $N_D$ giving an output image $O$ of width $w_D$ and height $h_D$ as:
\begin{equation}
O = N_D(I) 
\end{equation}
where
\begin{equation}
O(r\times h_D, (\theta / 2\pi) \times w_D) = I(x(r, \theta), y(r, \theta))
\end{equation}
and
\begin{eqnarray}
\label{eq:daugman1}\nonumber
x(r, \theta) = (1-r)\times (x_p + r_p\times cos(\theta)) + r\times (x_i + r_i\times cos(\theta))\\\nonumber
\label{eq:daugman2}
y(r, \theta) = (1-r)\times (y_p + r_p\times sin(\theta)) + r\times (y_i + r_i\times sin(\theta))
\end{eqnarray}
with $r\in[0,1]$ and $\theta\in[0, 2\pi]$. This linear model ``unrolls'' the iris region onto a rectangular region of a specified width and height.

In the next step, we utilize the iris domain-specific human-sources filters $F_{iris}$ \cite{czajka2019domain} that were learned using image patches extracted from human salient regions obtained from an eye-tracking device. We decided to use this iris feature extractor due to reported best recognition accuracy across all open-source iris matchers by a couple of teams \cite{Barni_Access_2021,Labati_IVC_2021}. Our identity preservation loss component is defined as:
\begin{equation}
L_{identity} = | F_{iris} \circledast N_D(I_O) - F_{iris} \circledast N_D(I_T) |
\end{equation}

\noindent
where $\circledast$ denotes the convolution operation. In words, we convolve the filters with the output iris images $I_O$ from the autoencoder, and target iris images $I_T$ after iris texture normalization $N_D$, and take the mean absolute error between these results as the identity-preserving loss.


\subsection{Maintaining Realism}
To match the shape of the generated iris with the desired shape (given by the input mask), we use the lightweight CC-Net~\cite{mishra2019cc} model, trained for the iris segmentation task, to find the logits for both of the images, and then minimize the absolute difference between the logits (``Mask Loss'' in Fig. \ref{fig:train}). 
This construction is differentiable and fast, so this enables us to use it in the autoencoder loss function without a significant slowdown in training. 

To ensure visual realism in the output images, we use the Learned Perceptual Image Patch Similarity (LPIPS-v0.1) loss (``Perceptual Loss'' in Fig. \ref{fig:train}), which matches deep features extracted from the AlexNet backbone to ensure the output image is ``perceptually'' similar to the target image ~\cite{zhang2018perceptual}. We also add the L1 norm between the generated and the ground truth samples (``L1 Loss'' in Fig. \ref{fig:train}), which attempts to match the output and target image directly. Ideally, if we had infinite data and complex-enough model architecture, this ``L1 loss'' would have been sufficient to effectively train our model. Practically, however, we need to build a mechanism into the overall training strategy that guides the network toward salient iris features in cases when data is limited. \rev{Note that the blurry outputs from the model are well applicable in biometrics, as iris recognition methods extract identity-related features within low spatial frequencies.}


\subsection{Choosing a Neural Network Architecture}

\rev{This section briefly summarizes a variety of experiments and architectures used prior to ending up with a successful model proposed in this paper. It may provide cues to other researchers about which architectures have the potential to work well in similar tasks.}

Since it is currently one of the most popular off-the-shelf generative adversarial networks (GANs), we initially experimented with the StyleGAN3 \cite{karras2021sg3} generator as a decoder to generate the output iris images. The generator model was trained from scratch on a subset of the WBPD dataset at a resolution of $512\times512$ pixels. Before initiating the generator training process, the data pre-processing included center-cropping the raw images to a square shape ($512\times512$), as well as removing raw images wherein the iris was located too close to the image boundary. Additionally, we filtered the training data to exclude iris images where little to no iris was present, such as when a subject was blinking. Left-right mirror augmentation was enabled during training, which effectively doubled the training data size. After training the generator and getting visually pleasing iris images, we took the generator, froze its weights, and trained an encoder to find the appropriate latent space representations that can preserve the identity. To train this encoder, we followed the training procedure as shown in Figure~\ref{fig:train}, however without the perceptual loss component that controls the visual reealism, as our belief was that GAN's generator can already produce visually appealing results (so this loss component would be redundant). One thing to note is that the autoencoder architecture that resulted from this did not contain skip connections as shown in Figure~\ref{fig:train}. 

Surprisingly, this intuitive approach ended up with disappointing results, which are still interesting to share as a ``negative'' yet useful information. Namely, we found that learning to find the latent space representation using an encoder in this way 
can neither preserve the identity nor produce visually-acceptable iris images. 
\rev{Finally, 
we found that U-Net architectures~\cite{isola2017image} provide better results than the StyleGAN3 generator, and experimented with different types of U-Net \ie UNet with ResNet connections, UNet with DenseNet connections, and UNet++. We found that more complicated architectures than the one provided in the paper produce similar results.}

\subsection{Architecture Details}
We utilize an autoencoder architecture with skip connections belonging to the family of U-Net architectures. Most U-Net architectures are optimized for image segmentation. Thus, to make U-Net ``better'' at generating iris images, not only segmentation results, we propose changes in the downsampling and upsampling operations. 

For downsampling, we replace max pooling in the original U-Net autoencoder with a combination of strided convolution and bilinear downsampling followed by convolution in parallel. In max pooling, gradients flow through only the max point and it works well when features are sparse. For our problem, we are looking to capture the overall iris texture and its dynamics, hence the information is not sparse. Our idea is that through strided convolution, the model can learn the most important features it wants to preserve, while the downsampling through interpolation would provide the model with a view of the overall features as well.

\begin{figure}[!ht]
\centering
\includegraphics[width=\linewidth]{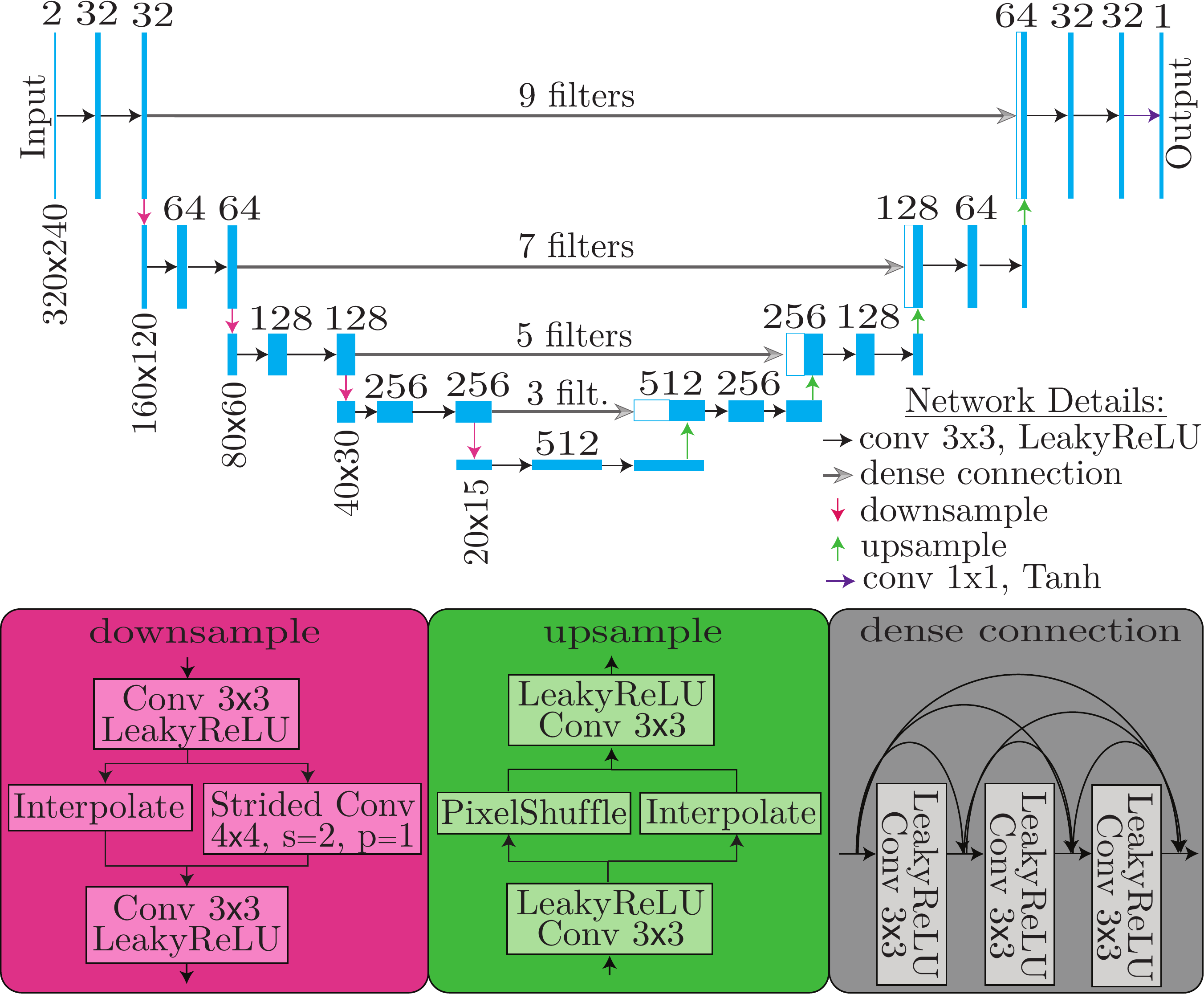}
\caption{Details of the proposed model and its modules. The bottom figure shows the detailed implementation of connections pictured in the top figure. The dense connections (illustrated in the bottom right picture) are explained for three filters; the same logic follows for the other number of filters.
}
\label{fig:network}
\end{figure}

For upsampling, we add bilinear upsampling followed by convolution in parallel to upsampling by sub-pixel convolution~\cite{shi2016real, aitken2017checkerboard}. Using a transposed convolution, as in the original U-Net, has been known~\cite{odena2016deconvolution} to cause checkerboard artifacts in the produced image, especially when the kernel size is not divisible by the stride causing uneven overlap of the kernel while sliding over the image. While the transposed convolution used in the original U-Net architecture had a kernel size and stride of 2, we found that replacing it as detailed decreases artifacts in the produced image, and thus improves the performance of the iris recognition applied to images generated by this model. Figure~\ref{fig:network} shows the overall architecture of our network.

\section{Applications}

There are two biometric applications of the proposed model. First, a model capable to generate ISO-compliant iris images with varying pupil sizes, and deforming the iris texture as the authentic eye does, should increase the accuracy of any iris matcher by comparing images with rectified pupil size, instead of raw iris scans. Second, visually-realistic and identity-preserving iris images of varying pupil sizes can increase the accuracy of human examination in forensics. We present details of these two applications in the following subsections.

\subsection{Iris Recognition}
The usefulness of the proposed model in iris recognition is evaluated on the subject-disjoint test set of Warsaw-BioBase-Pupil-Dynamics v3.0~\cite{kinnison2019learning} (WBPD). The example iris recognition method utilized here is based on human-driven binary image features ~\cite{czajka2019domain}. We compare our deformation methodology with the linear approach~\cite{daugman1993high} (denoted later as ``Linear'') and the nonlinear biomechanical model~\cite{tomeo2015biomechanical} (denoted later as ``Biomech''). 

As there can be a huge number of pairs if we make all possible comparisons for the WBPD dataset, we randomly select image pairs from the minimum pupil-to-iris ratio (0.1 to 0.2) bin and maximum pupil-to-iris ratio (0.7 to 0.8) bin without replacement for each individual. We find genuine and imposter comparison scores using these pairs. As we are selecting the pairs randomly, we repeat the experiment 10 times to assess the statistical significance of the results.


\begin{table}[!ht]
\caption{Average percentage of different iris code bits for genuine and imposter comparisons and for various iris texture deformation models.}\vskip2mm
\label{table:bitdiff}
\centering
\begin{tabular}{l|l|l}
{\bf Method} & {\bf Genuine } & {\bf Imposter } \\
& {\bf  pairs} & {\bf  pairs} \\ \hline\hline
Linear \cite{daugman1993high} & 32.67 $\pm$ 0.22 & 43.81 $\pm$ 0.15 \\ \hline 
Biomech \cite{tomeo2015biomechanical} & 32.74 $\pm$ 0.21 & 43.78 $\pm$ 0.16 \\ \hline
DeformIrisNet (proposed) & 22.40 $\pm$ 0.11 & 34.02 $\pm$ 0.08 \\ \hline
\end{tabular}
\end{table}

The iris recognition method used in this evaluation 
provides unique binary codes for the iris textures of different individuals. Table~\ref{table:bitdiff} reports the percentage of bits that are different for the binary codes between authentic gallery images (with smaller pupil) and rectified probe images (with the pupil size matching the size of the authentic gallery samples) for all transformation methods. 
As we can see from the table, the deformed pupil image generated by DeformIrisNet has a lower number of disagreeing bits between the authentic gallery and rectified genuine probe image than in the case of other methods. \rev{This is good and expected as it indicates that deformed iris images are now closer, in terms of the features utilized by the iris recognition module, to authentic images. However, we also observe a lower number of disagreeing bits for imposter comparisons, which is not desired but likely happens because we are deforming the two iris images from different individuals to the same iris shape.}
Thus, to evaluate the efficacy of the DeformIrisNet model in terms of the biometric identification capabilities when used in conjunction with an iris recognition module, we find the genuine and imposter score distributions for all the iris normalization methods and calculate the decidability ($d'$) defined as:
\begin{equation}
    d' = \frac{|\mu_g - \mu_i|}{\sqrt{\frac{1}{2}(\sigma_g^2 + \sigma_i^2)}}
\end{equation}
where $\mu$ and $\sigma$ are the mean and standard deviation, respectively, of the genuine ($g$) and imposter ($i$) scores. We also calculate the Equal Error Rate (EER) for all the methods. 

\begin{table}[!ht]
\caption{Decidability score $d'$ and Equal Error Rates (EER) obtained in {\bf same-dataset subject-disjoint evaluation}.}\vskip2mm
\label{table:dprime}
\centering
\begin{tabular}{l|l|l}
{\bf Method} & $d'$  & {\bf EER} \\ \hline\hline
Linear \cite{daugman1993high} & 2.642 $\pm$ 0.072 & 0.121 \\ \hline
Biomech \cite{tomeo2015biomechanical} & 2.644 $\pm$ 0.062 & 0.115 \\\hline
DeformIrisNet (proposed) & 3.003 $\pm$ 0.049 & 0.118 \\ \hline
\end{tabular}
\end{table}



\begin{figure}[!ht]
\centering
\includegraphics[width=\linewidth]{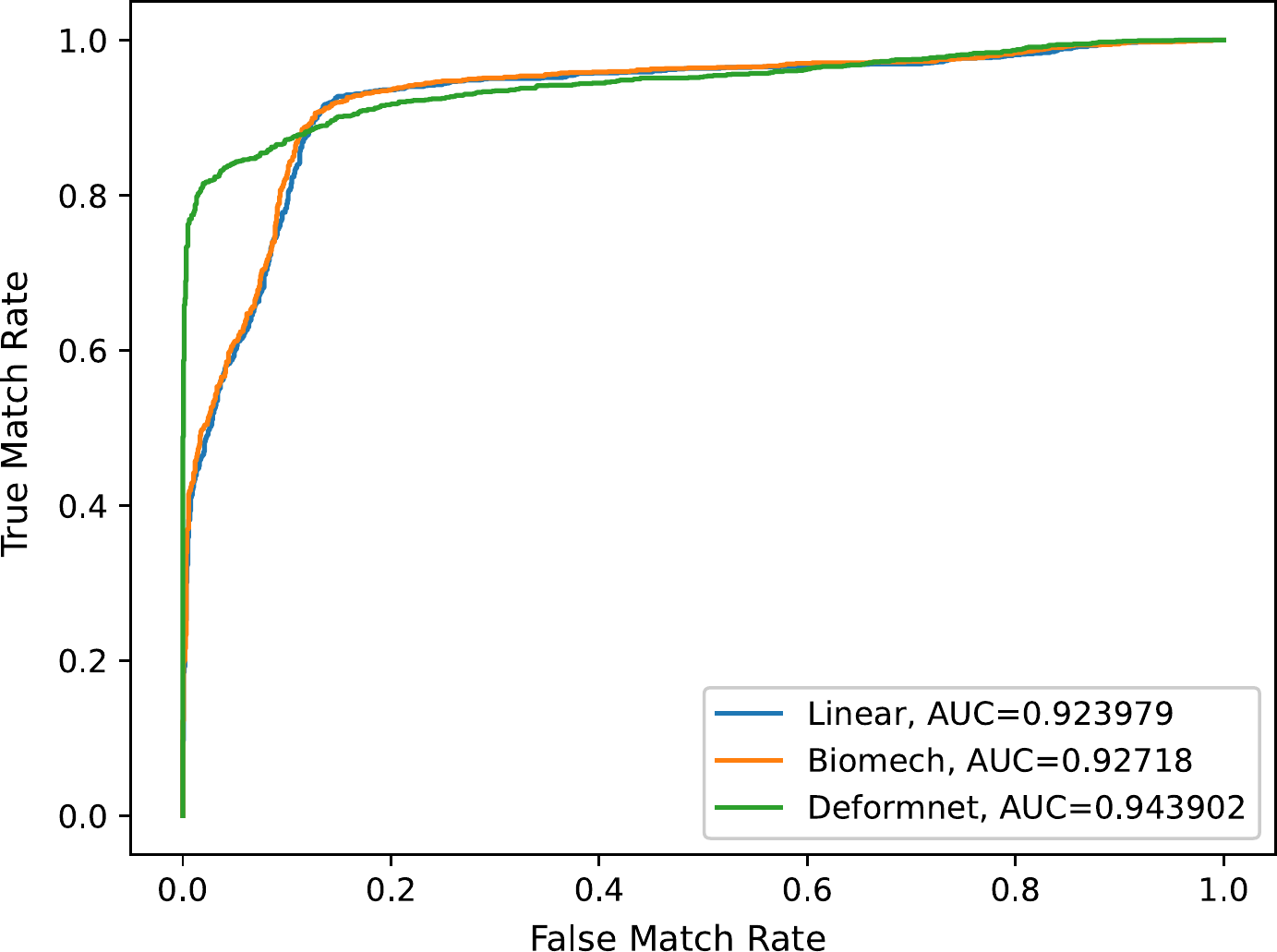}
\caption{ROC curve and Area Under the ROC Curve values for iris recognition with different iris texture deformation methods: Linear~\cite{daugman1993high}, Biomech~\cite{tomeo2015biomechanical}, and DeformIrisNet (proposed)}
\label{fig:roc_curve}
\end{figure}

\rev{As shown in Table~\ref{table:dprime}, we find that using the DeformIrisNet model a higher separation between the imposter and genuine score distributions (larger $d'$) can be achieved, indicating that our model is able to produce rectified iris images that can improve iris recognition results compared to those obtained by linear or biomechanical-based normalization strategies.}  While EER is on par with other methods, the receiver operating characteristic (ROC) curves shown in Figure~\ref{fig:roc_curve} suggest that the example iris recognition method when applied to images rectified with the proposed nonlinear iris deformation model, obtains a better recognition accuracy, measured by the Area Under the ROC Curve.

\subsection{Human Examination}

With iris recognition becoming the next biometric modality in e-passports, a component of the FBI's Next Generation Identification (NGI) system \cite{NGI}, and recently documented its usefulness for identification of deceased subjects \cite{Trokielewicz_TIFS_2019}, the need for having trained (professional) human iris examiners, who could confirm the machine's decision in a legally-biding manner started to emerge. For instance, NIST has initiated a working group that meets regularly and aims at designing a training curriculum for human iris image examiners who might be called upon to give testimony in court \cite{IEG-Examiners-Group}. 

\begin{figure}[!h]
\centering
\includegraphics[width=\linewidth]{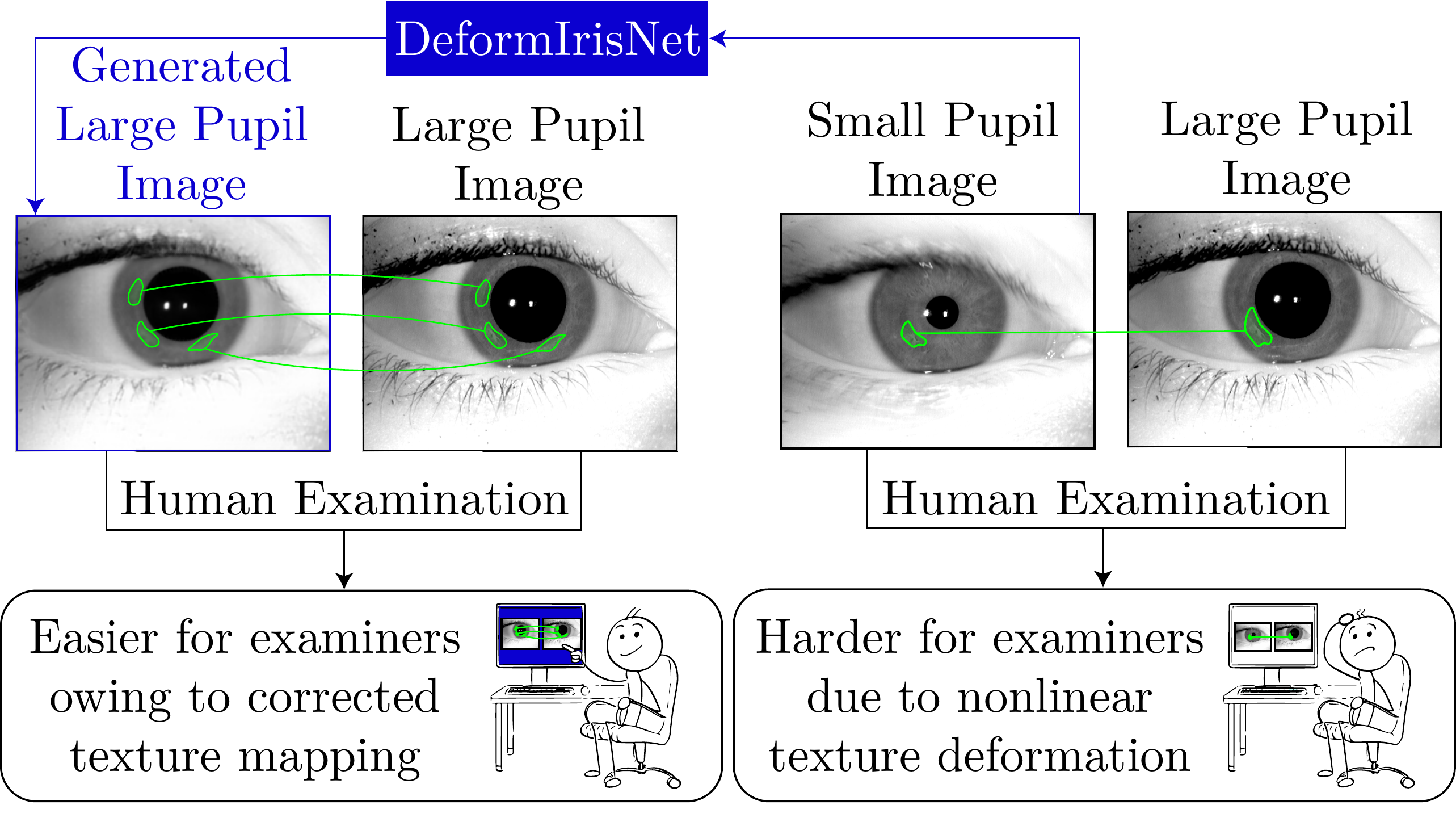}
\caption{Comparing iris images with the same pupil size is much easier for human examiners than samples with an excessive difference in pupil size. DeformIrisNet can be used to rectify the size of the iris annulus, applying correct, learned from the data nonlinear warping of iris texture.}
\label{fig:humanexam}
\end{figure}

\begin{figure*}[!ht]
\centering
\includegraphics[width=0.93\textwidth]{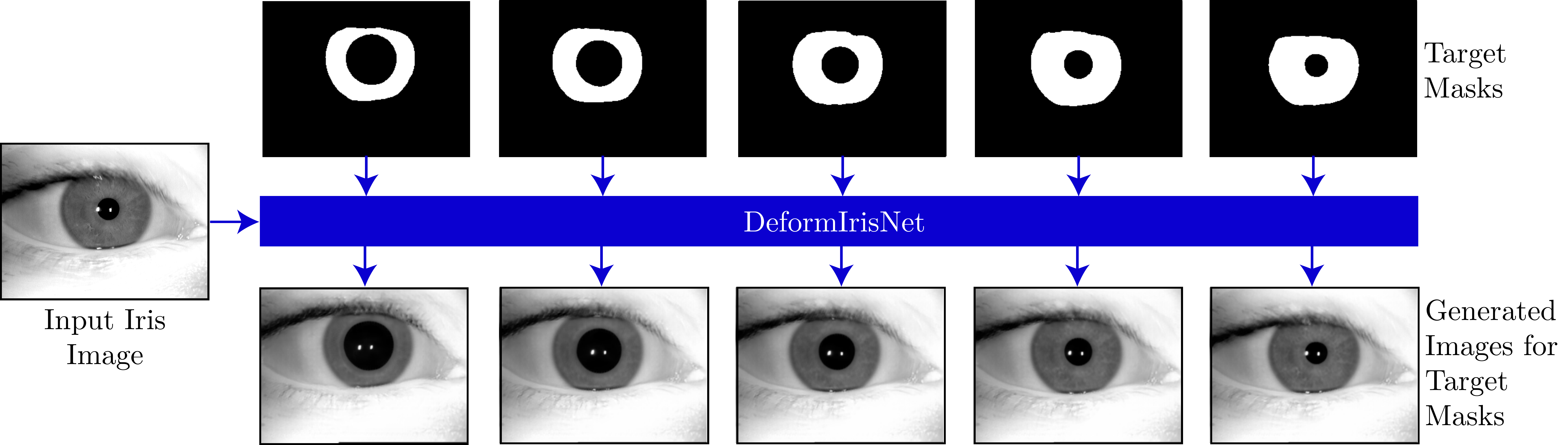}
\caption{Iris images generated by DeformIrisNet from a single iris sample (left), given the target masks (top) that can be used in human examination of iris image pairs, making the process easier for forensic examiners when examined iris scans with varying pupil size.}
\label{fig:samples_fidm}
\end{figure*}

\begin{figure*}[!ht]
\centering
\includegraphics[width=0.93\textwidth]{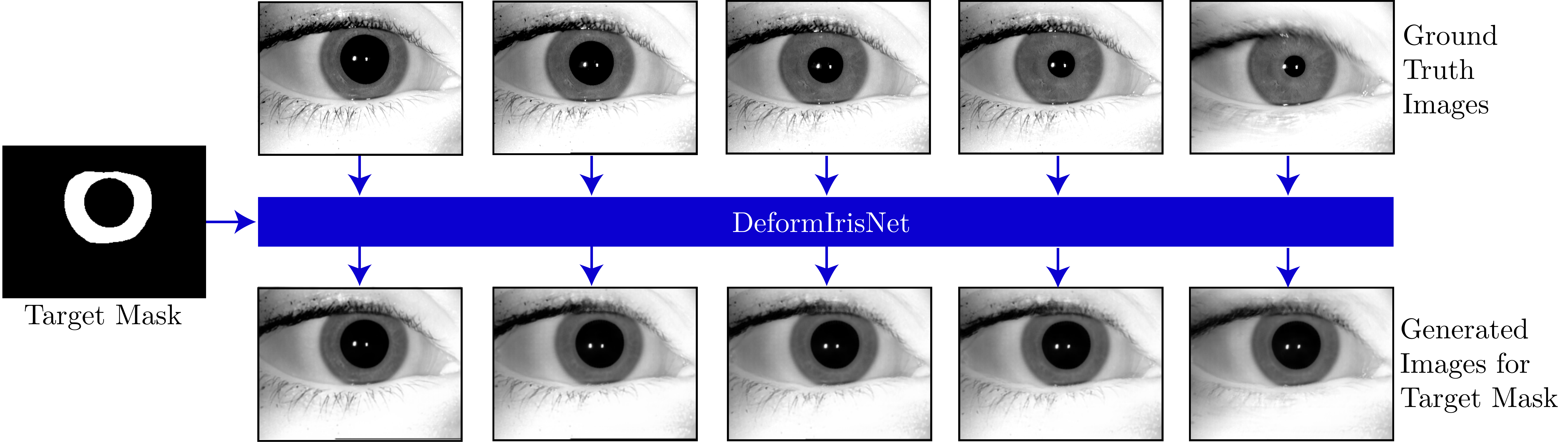}
\caption{Same as in Fig. \ref{fig:samples_fidm}, except that illustrating generative process from multiple images of varying pupil size into a single iris image with canonical pupil size.}
\label{fig:samples_difm}
\end{figure*}

The proposed deformation model may become a useful part of the human examination toolbox. As demonstrated in the literature \cite{Moreira_WACV_2019} and illustrated in Figure~\ref{fig:humanexam}, human examiners may have a tough time comparing iris images from the same individual with an excessive difference in pupil size in both samples. Our iris deformation model allows us to generate iris images with varying pupil sizes from a still iris image. That is, the human examiner, by rotating a virtual ``knob'', can generate an infinite number of samples preserving the identity and showing the iris at various levels of constriction to match the desired pupil size. The project GitHub repository\footnote{\url{https://github.com/CVRL/DeformIrisNet}} includes a video showing an example output of this process, illustrated also in Fig. \ref{fig:samples_fidm}. Such a tool should significantly increase the chances of correctly matching iris samples under human examination, and to our knowledge has never been proposed before.



Figure~\ref{fig:samples_difm} additionally shows the result of running multiple iris images with different pupil sizes through DeformIrisNet and requesting the same pupil size for all samples. An interesting observation here is that the smaller the change in the pupil is, the sharper and closer the output image is to the input image. This should be the case because the smaller the difference in the pupil size, the ``lesser'' the work that our model needs to do to deform the iris image and match the input pupil mask. \rev{For human iris examination (which is not the focus of this paper), sharper images (showing more fine-grained iris texture) would be more desirable. However, to quantitatively assess this statement, human examination experiments, in which subjects compare iris images with and without proposed normalization, are needed.}

\section{Iris Inpainting and Noise Reduction}

\rev{Apart from two core applications driven by actual and timely needs in the iris recognition community, the proposed model can be also used for ``painting'' realistically-looking iris NIR images given an arbitrary iris mask. Also, it can be applied as an iris domain-specific noise reduction tool. However, these are ``by-products'' of our model and our training mechanism did not optimize the model for either of these tasks. This subsection discusses these observations.}

\begin{figure}[!ht]
\centering
\includegraphics[width=\linewidth]{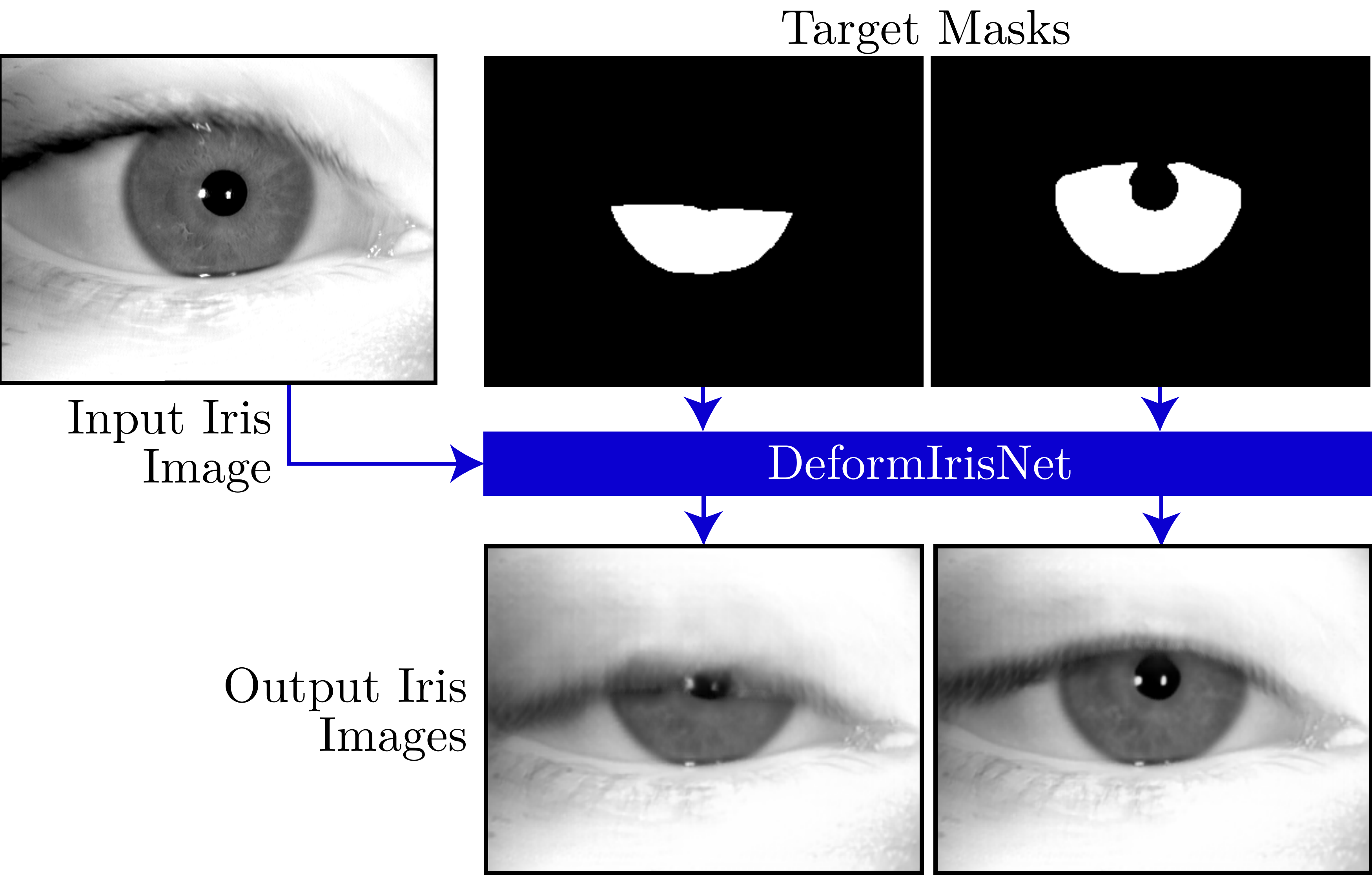}
\caption{Generating partially closed eye images from an open eye image given the segmentation mask.}
\label{fig:open2closed}
\end{figure}

In the training dataset we utilized, there are iris images with partially closed eyelids when the subject was blinking. Interestingly, seeing such examples during training, the model learned how to ``open'' or ``close'' the eyelids given the mask suggesting such situations. Figure~\ref{fig:open2closed} shows what happens when we take an iris image and run it through the model together with masks for partially closed-eye images. We see that the network effectively learned to close the eyelid while preserving the iris texture for the visible region.

\begin{figure}[!ht]
\centering
\includegraphics[width=\linewidth]{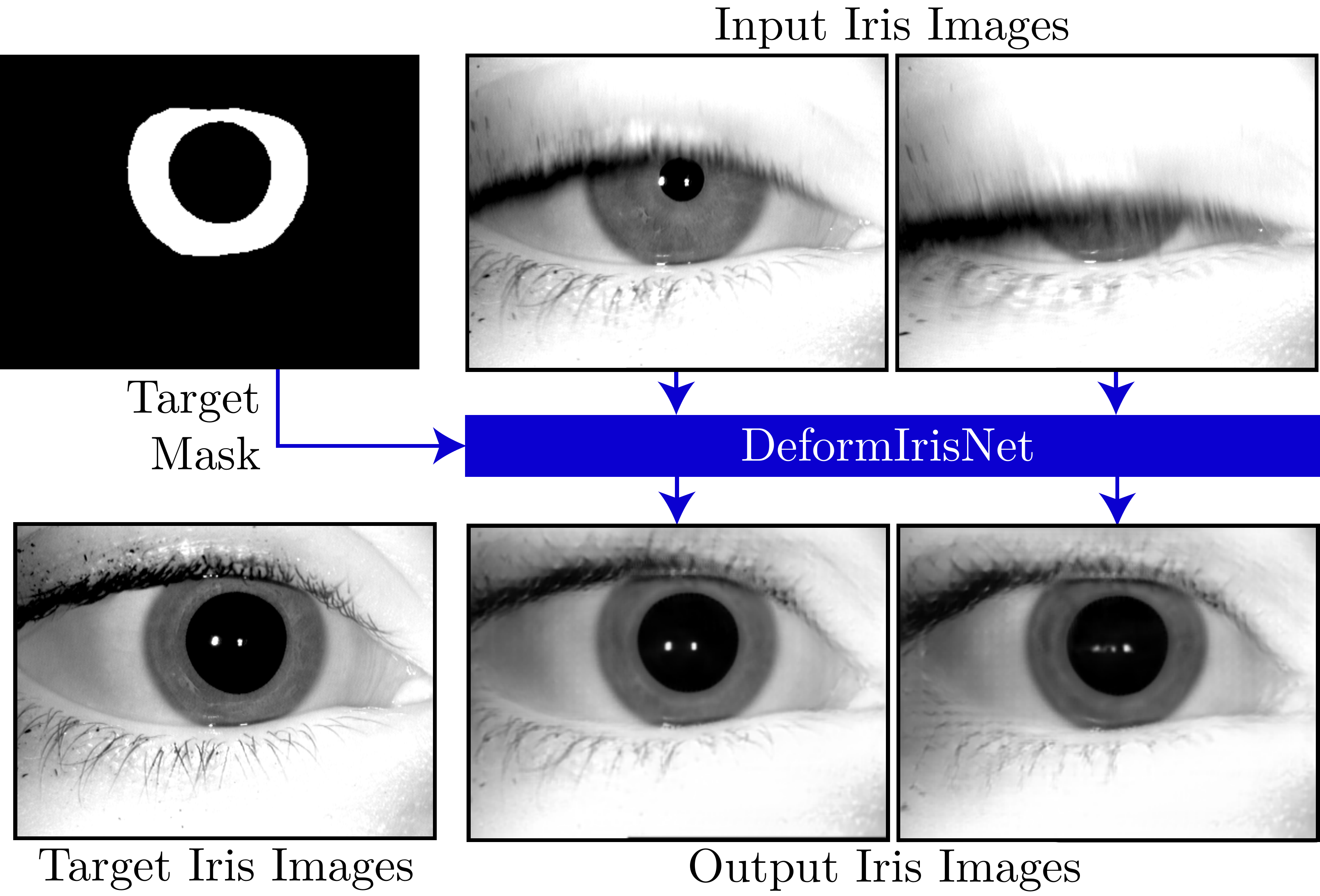}
\caption{Generating an open-eye image from a partially-closed-eye image given the segmentation mask.}
\label{fig:closed2open}
\end{figure}

Figure~\ref{fig:closed2open} shows what happens when we take an iris image with partially closed eyelids and run it through the model with a target mask suggesting a fully opened eye. Interestingly, if a significant portion of the iris texture is visible, the rest of the iris texture is being ``dreamed'' up by the network in a way to make it similar to the actual iris texture, but -- certainly -- without any guarantee that the generated texture represents a given identity, \rev{hence having limited application in biometric recognition.}

\begin{figure}[!ht]
\centering
\includegraphics[width=\linewidth]{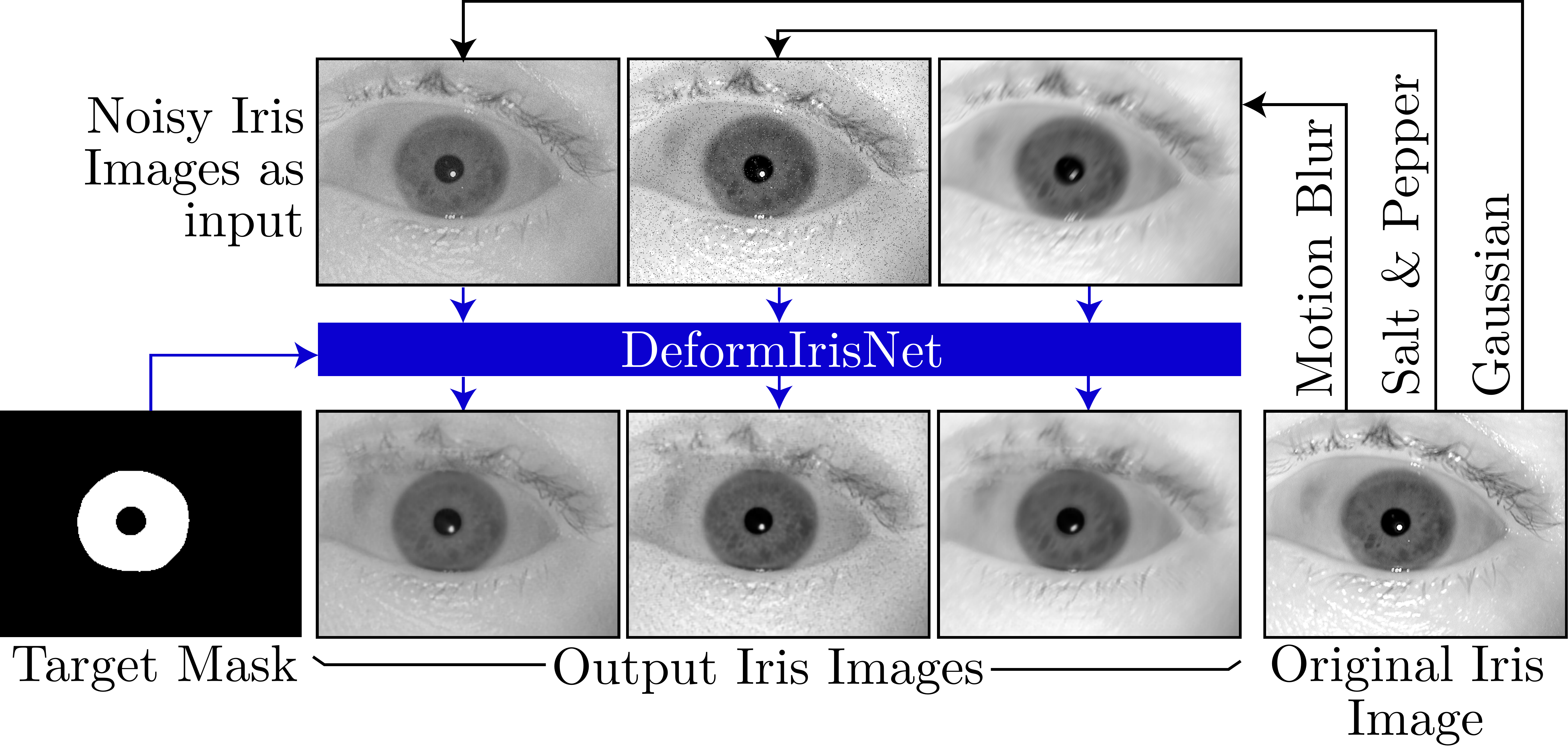}
\caption{An example of what happens when we run a noisy version of the image through the network.}
\label{fig:effectnoise}
\end{figure}

Finally, Figure~\ref{fig:effectnoise} demonstrates how the trained DeformIrisNet can partially de-noise iris images by feeding it with a noisy iris image and actual segmentation mask for that image (so no changes in iris texture are requested). While it is not perceived as a main application of the trained model, we found it an interesting by-product for the model that ``understands'' the complex deformations of the iris texture. \rev{We believe training specifically for this ``denoising'' task can provide interesting results~\cite{gondara2016medical, jifara2019medical}.}

\section{Conclusions}
This paper proposes an end-to-end, fully data-driven autoencoder-based approach to mimic complex deformations of the iris texture while preserving the identity information of an individual whose iris image is being processed. \rev{Our method aims to mitigate the degradation in performance that occurs in iris recognition modules due to changes in pupil size by filling the gap between the accuracy observed for comparing irises with same-size and different-size pupils. We demonstrated its potential usefulness in iris recognition and propound its usefulness in forensic human examination, since matching the iris images with identical pupil size should make it easier for human examiners to compare and match iris scans.} Since the rectified iris images are compliant with ISO requirements, they can be used with any iris recognition methods, including black-box / closed-source commercial solutions. In addition, we presented the in-painting and noise reduction capabilities of the proposed model. Source codes and model weights are offered along with the paper for full reproducibility. 

{\small
\bibliographystyle{ieee}
\bibliography{main}
}

\end{document}